\documentclass[10pt, conference]{IEEEtran}
\IEEEoverridecommandlockouts
\usepackage{graphicx}
\graphicspath{/Users/tunderwood/Dropbox/Documents/Active research/BigHumanities/}
\DeclareGraphicsExtensions{.pdf,.eps}
\usepackage{booktabs}
\begin{document}
\title{Mapping Mutable Genres \hspace{\textwidth} in Structurally Complex Volumes}

\author{\IEEEauthorblockN{Ted Underwood, Michael L. Black}
\IEEEauthorblockA{Department of English\\
University of Illinois, Urbana-Champaign\\
Urbana, IL, USA\\
tunder@illinois.edu, black7@illinois.edu}
\and
\IEEEauthorblockN{Loretta Auvil, Boris Capitanu}
\IEEEauthorblockA{Illinois Informatics Institute\\
University of Illinois, Urbana-Champaign\\
Urbana, IL, USA\\
lauvil@illinois.edu, capitanu@illinois.edu}
}
\IEEEpubid{\makebox[\columnwidth]{\hfill Accepted for the 2013 IEEE International Conference}
\hspace{\columnsep}\makebox[\columnwidth]{on Big Data, Oct 6-9, Santa Clara, CA, USA. \copyright~2013 IEEE \hfill}}
\maketitle

\maketitle

\begin{abstract}
To mine large digital libraries in humanistically meaningful ways, we need to divide them by genre. This is a task that classification algorithms are well suited to assist, but they need adjustment to address the specific challenges of this domain. Digital libraries pose two problems of scale not usually found in the article datasets used to test these algorithms. 1) Because libraries span several centuries, the genres being identified may change gradually across the time axis. 2) Because volumes are much longer than articles, they tend to be internally heterogeneous, and the classification task also requires segmentation. We describe a multilayered solution that trains hidden Markov models to segment volumes, and uses ensembles of overlapping classifiers to address historical change. We demonstrate this on a collection of 469,200 volumes drawn from HathiTrust Digital Library.
\end{abstract}

\section{Introduction}
Many attempts to mine large historical collections have treated them as a single pool of documents differentiated only by publication date [1, 2]. While some of this work has been groundbreaking, humanists have often expressed skepticism about the underlying assumption that culture changes as a unified whole. For instance, the increasing frequency of the word ``toddler'' in Google Books has been offered as evidence that postmodern novelists had a soft spot for children [3]. But if ``toddler'' became more common because more parenting manuals were published, this evidence might have no relation at all to the history of the novel.

Before text mining can make a real contribution to the humanities, then, we need a way to divide large collections into meaningful subsets. In practice, humanists are especially interested in categories we call ``genres'' --- novels, for instance, or lyric poems, or sermons. Unfortunately, metadata is not readily available to classify volumes by genre. Librarians estimate that genre information is present in the expected MARC field for less than a quarter of the volumes in HathiTrust Digital Library [4]. Moreover, even if we had complete information at the volume level, we would still confront the deeper problem that volumes are internally heterogeneous. A nineteenth-century novel often begins with a nonfiction life of the author, and ends with twenty pages of publishers' ads. To distinguish the history of the novel from the history of its paratext, we need some way of separating genres inside a volume. 

A great deal of work has already been done on classification by genre: it's a problem that has preoccupied linguists and information scientists as well as critics since Aristotle. But no consensus has emerged, and many scholars now suggest that genre may lack a stable definition [5, 6]. At one end of the spectrum, a genre can be a clearly-defined form like a sonnet or an index. But the word also embraces formally protean discourses like satire and science fiction, which have no fixed structure. In practice, writers use \emph{genre} to describe any rhetorical situation that isn't wholly reducible to a topic. But this need not imply that genre and topic are perpendicular axes of classification: some genres, like the travel guide, are defined largely by subject matter.

Fortunately, it's not necessary to confine this flexible concept in order to support humanistic research. Our task is not to redefine genre, but simply to sort texts into categories that researchers already in practice find socially meaningful. The instability of the underlying concept is important mainly because it warns us not to expect that genre will reside in any specifiable dimension of language (e.g. in form rather than content, or syntax rather than semantics).

In practice, many genres can be identified using the same bag-of-words model that supports other forms of text classification [7, 8]. To adapt this model for genre, it's only necessary to put back some of the things that researchers take out when they want to focus on subjects. For instance, we find that it's not a good idea to lemmatize words or remove stopwords: verb tenses and prepositions provide significant clues. But these are minor adjustments. At its core, genre classification can be a straightforward application of learning algorithms that are known to work well on text. Naive Bayes and regularized logistic regression both produce good results.

The problem becomes more interesting, however, as we move away from its algorithmic core and toward specific kinds of scale that complicate the humanistic domain. 
\IEEEpubidadjcol

\section{Historical heterogeneity.}

The first of these complications involves time: in a dataset that spans centuries, genres can't be treated as mutually exclusive categories, or as fixed points of reference. 

For instance, suppose we want to identify book-length works of prose fiction published in English in the eighteenth and nineteenth centuries. HathiTrust Digital Library can provide a digital collection of 469,200 nonserial English-language volumes from that period, along with metadata from the member libraries. This is a good place to start, but selecting works of fiction from this collection is challenging. Existing metadata rarely provides unambiguous information about genre. More troublingly, when you dig into the problem, it becomes clear that no amount of manual categorization will ever produce a definitive boundary between fiction and nonfiction in a collection with a significant timespan, because the boundary changes over time. Form and content didn't necessarily align in earlier centuries as they do for us. Nineteenth-century biographies that invent imagined dialogue often read exactly like a novel; eighteenth-century essays like Richard Steele's \emph{Tatler} use thinly fictionalized characters as a veil for nonfiction journalism.

If fiction and nonfiction are hard to separate, subtler generic categories like fantasy and science fiction will be even harder. So a study of genre has to begin by acknowledging that genres blur and overlap. Oddly enough, this becomes easier to acknowledge if you represent genre quantitatively. Although we like to mock computers' reliance on ``binary" logic, the point of using numbers is after all to represent questions of degree. The discipline of machine learning would call genre a problem of \emph{multi-label} classification: a given work can belong to any number of genres to different degrees [9]. Those degrees of membership needn't always operate under a zero-sum logic. Some categories, like fiction and nonfiction, probably do trade off against each other in zero-sum fashion, but a novel doesn't necessarily become less like ``gothic'' as it becomes more like ``romance.''

Classification algorithms adapt very naturally to these shades of gray, expressing their predictions as degrees of probability. Moreover, because algorithmic classification doesn't require a great deal of human labor, it can be undertaken in a provisional and exploratory way. To be sure, there are aspects of genre that algorithms based on word frequency will fail to capture. For instance, they're not good at discerning cases where a work instantiates a genre in order to parody it. Some genres may not be recognizable at all with a bag-of-words model. But for broad distinctions between fiction, drama, poetry, and nonfiction prose, classification algorithms can be a powerful and surprisingly flexible mapping tool.

On the other hand, in pursuing this approach it's vital not to lose sight of humanistic historicism. Social categories like fiction don't have constant definitions [10]. As we've seen, the boundary separating fiction from nonfiction can change over time. Moreover, language itself changes. Some of the linguistic features that characterize fiction remain constant from the eighteenth century through the nineteenth: past-tense verbs of speech and personal pronouns are prominent. But the particular assortment varies: eighteenth-century characters do a lot of replying, where nineteenth-century characters interrupt and exclaim.

\begin{figure}[!t]
\centering
\includegraphics[width=\linewidth]{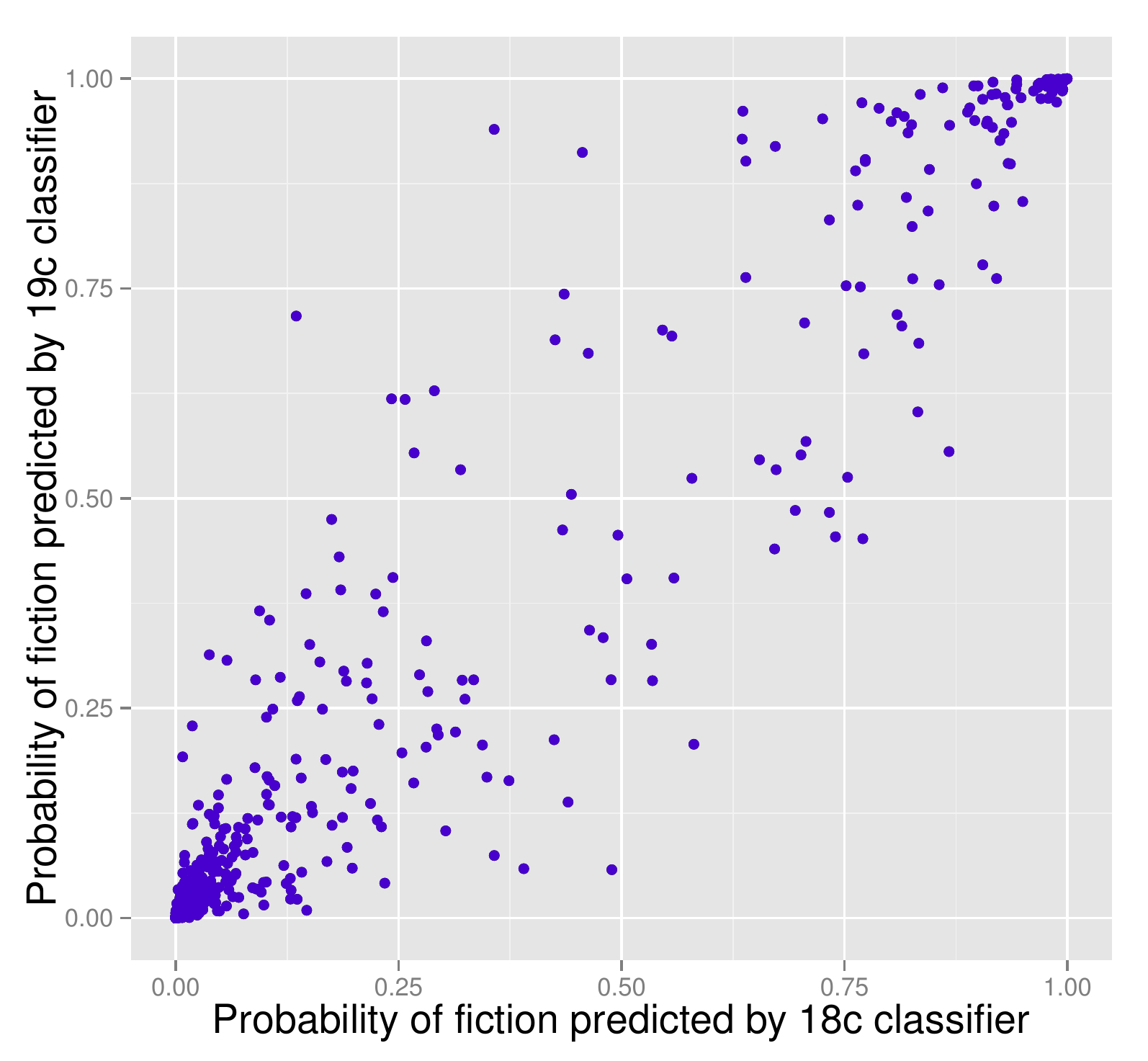}
\caption{Scatterplot of 1000 volumes randomly selected from a corpus of 469,200 18th- and 19th-century volumes. Classifiers trained on different centuries predicted the probability that each volume was fiction.}
\label{fig_class}
\end{figure}

One simple way to acknowledge the mutability of language and of genre is to train multiple classifiers with evidence drawn from different slices of time. Then we can use publication dates to weight their votes on the probability that a given document belongs to a given genre. In some cases, as Fig. 1 indicates, there will be a great deal of agreement between different classifiers. Here we've trained two naive Bayes classifiers on collections, respectively, of eighteenth- and nineteenth-century volumes. We categorized volumes manually to distinguish prose fiction, prose nonfiction, drama, and nondramatic poetry. The classifers learned to model fiction by observing the differences that separated it from other genres; they were then calibrated to express their predictions probabilistically, using a logistic rather than linear function---which is why there are more dots in the corners of the graph than in the middle. (A logistic function changes its value more rapidly in the middle of its range than at the ends.)

In this case there was in practice a strong positive correlation (0.963) between models trained on evidence from different centuries. But this won't always be true. In attempting to recognize gothic novels, we've found it difficult to train a single classifier that will recognize both Romantic-era gothic (e.g., \emph{Frankenstein}) and late-nineteenth-century gothic (e.g., \emph{Dracula}). If we want to model gothic as a single genre, we'll have to do it with a chain of overlapping but significantly different models spread out over time. Of course, the continuity of gothic is actually a debatable premise: some critics would call gothic a ``mode," like tragedy, rather than a coherent genre, like detective fiction. Statistical modeling can't confirm or deny a premise like this single-handedly, but it can resist the assumptions we bring to it, by showing that on the level of diction at least, gothic lacks the coherence of some other genres (e.g. the Robinsonade; see below).

To extract fiction from the HathiTrust collection, we actually used three overlapping classifiers (one for the eighteenth century, one for the nineteenth, and one for the period 1750-1850). We weighted their votes according to the date of each volume, and combined the evidence of diction with a cautious filter based on metadata. We ended up with a collection of 32,209 volumes, which may be conservative. The distribution strongly leans toward the end of the timespan, both because publication generally increases in the nineteenth century, and because the prominence of fiction increased steadily from 1700 (when it constituted less than 5\% of the HathiTrust collection) to 1899 (when it amounted to at least 20\%). Many of these volumes are reprints, or multiple parts of a single work, so the collection doesn't actually cover 30,000 distinct titles. We will eventually deduplicate the corpus. But for historical purposes it's often an advantage to leave the reprints in: an argument can be made that frequently-reprinted works should have more weight in our collection.

\subsection{An experiment: point of view.}

Once you've mapped genre in a collection, what can you do with it?

Many important aspects of literary history have been ignored because they're difficult to address at the scale of reading we ordinarily inhabit. For instance, the choice between first- and third-person points of view is one of the most fundamental decisions an author can make. But critics know surprisingly little about its history. Histories of fiction that focus on selected examples can do an excellent job of tracing formal innovations, because it makes sense in that case to focus on a small number of innovators. Accordingly, we know quite a lot about refinements of perspective like free indirect discourse. But the broad categories of first- and third-person narrative have been around for centuries: to describe the shifting balance between them would require evidence about proportions, rather than innovation as such. A few hundred canonical examples don't tell us much about proportions. Thousands of novels were published in the nineteenth century, and it seems quite likely that critics' preferences diverged from the preferences of the broader book-buying public. Library collections may also have diverged from the tastes of the book-buying public, but they give us at least a larger sample.

Fortunately, point of view can be mapped using the same techniques we use to map genre. It's rather easy to distinguish first- and third-person works of fiction using a bag-of-words model, because point of view markedly affects (among other things) the frequencies of personal pronouns. First-, second- and third-person pronouns are always present in dialogue, but narrative perspective alters their proportions in the work as a whole. Of course, the choice between first and third person isn't an all-or-nothing question: there are works (like Dickens' \emph{Bleak House}) where they alternate. They can also blur in less explicit ways: a first-person narrator can be so reticent as to almost disappear from his own narrative. But we need to stress once again that classification algorithms are comfortable mapping shades of gray. An algorithm can't intimately characterize the narrators of Jane Austen and Henry James, but it can recognize cases where the ideal types of first- and third-person perspective are mixed. Our algorithm in fact recognized that \emph{Bleak House} was a boundary case with nearly equal proportions of first- and third-person narration.

\subsection{Methods.}

We have suggested that recognition of rhetorical concepts like genre and point of view requires no special recipe. For the most part, we use supervised learning algorithms that are already known to work well in other contexts---naive Bayes and regularized logistic regression. On the other hand, libraries of volume-length documents do differ from the article datasets ordinarily used to test these algorithms in one simple but important way: the average volume is much longer than the average article.

This can introduce tricky problems of internal heterogeneity discussed at more length below. But in other respects it makes the task of classification easier. Text classifiers ordinarily have to grapple with very sparse data: there are thousands of potentially relevant words, but only a few hundred may actually be present in a given web page or article. Many of those words occur only once. For this reason classifiers sometimes use a Bernoulli model, ignoring the number of times a word occurs, and attending only to the binary question of its presence or absence in a document.

The volumes in HathiTrust average roughly 100,000 words long. This makes sparsity a much less significant problem. A Bernoulli model would be counterproductive here, because documents are distinguished less by their (strongly overlapping) vocabularies than by details of word frequency. It also becomes possible to do some other things that aren't usually attempted in text mining. 

For instance, corpus linguists have recommended a Wilcoxon ranksum test as the best way to identify words that characterize one corpus in contrast to another [11]. Instead of summing all occurrences of a word in both corpora, the Wilcoxon test considers documents individually to ensure that the word is \emph{consistently} more common in instances drawn from one corpus. This test is often used for selecting  features in bioinformatics [12], but rarely in text mining---perhaps because comparing word frequencies in short documents produces many ties or near-ties that are hard to interpret. Given this sparse data, a measure like mutual information might be more robust. But volume-length documents are more comparable to a genome: words are expressed at many different frequency levels, and it's often possible to rank their frequencies without ties.

We've used the Wilcoxon test heavily to identify sets of features for classification. For instance, in classifying point of view we started with information about the 1200 most common words in 288 training examples (divided between first- and third-person novels). But from that list of 1200 words, we allowed the Wilcoxon test to select 20 words that positively characterize first-person novels, and 20 words that are more common elsewhere. We trained a naive Bayes model on this reduced set of 40 features.

Tenfold cross-validation confirms that this is an effective model (see Table II). Other sets of features might work equally well, but we're interested in feature selection not only to increase accuracy but as a way of interpreting the lexical contrast between two corpora. For this purpose the Wilcoxon test has significant advantages: instead of representing the internal mechanics of classification (like ``information gain''), it's immediately legible as a recurring contrast between instances of two classes. In classifying novels by point of view, this method turned out to be interesting, for instance, because it drew our attention to the odd fact that first-person novels rely disproportionately not only on ``I" and ``me," but on numbers.

\begin{figure}[!t]
\centering
\includegraphics[width=\linewidth]{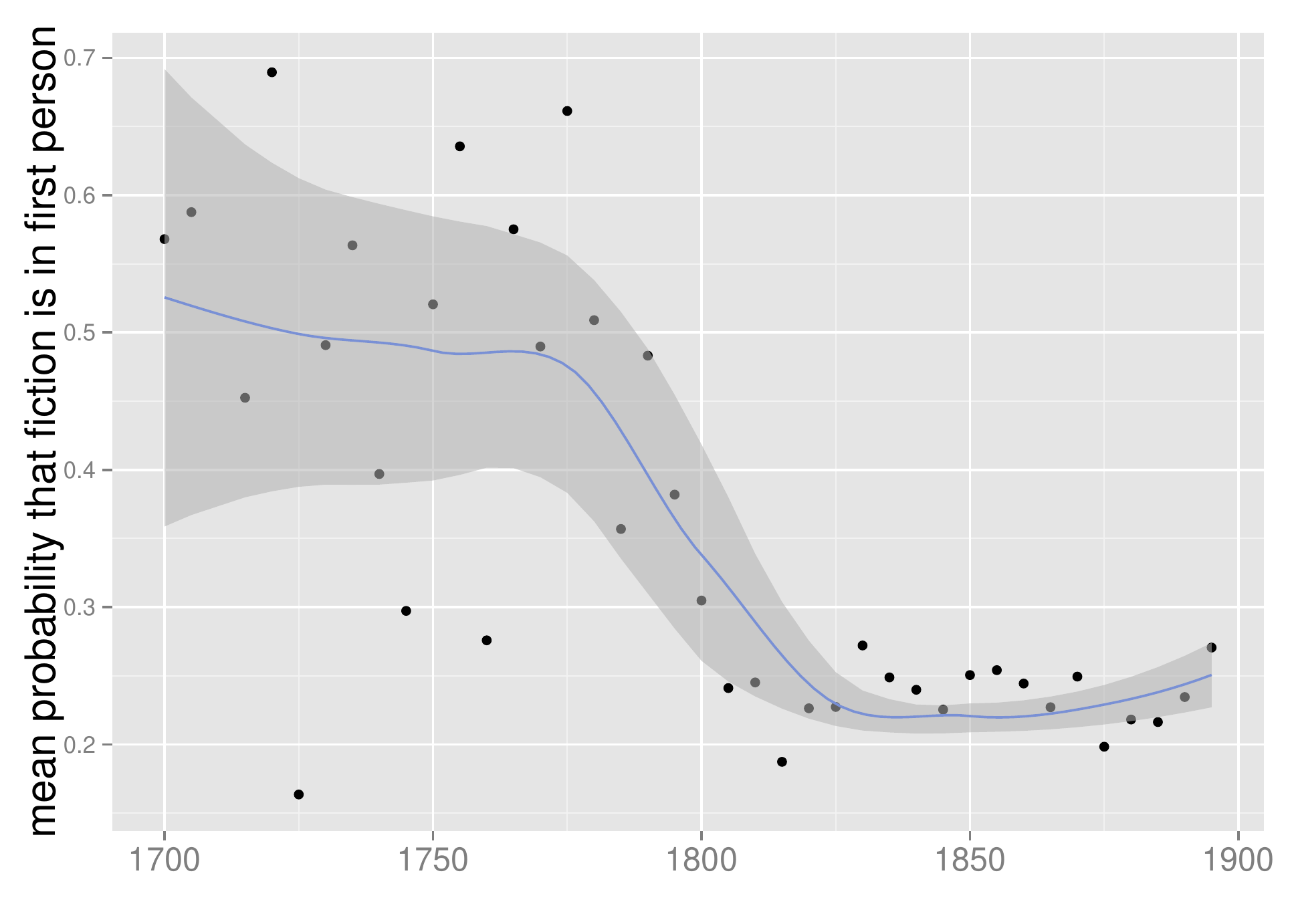}
\caption{Mean probability that fiction is written in first person, 1700-1899. Based on a corpus of 32,209 volumes of fiction extracted from HathiTrust Digital Library. Points are mean probabilities for five-year spans of time; a trend line with standard errors has been plotted with loess smoothing.}
\label{fig_class}
\end{figure}

\subsection{Interpreting the results.}
The simplest thing we discovered with this experiment was that the the proportion of fiction written in the first person declines significantly toward the end of the eighteenth century. Fig. 2 plots the mean probability of first-person narration at five-year intervals, and reveals a shift from a regime where first person narration is used roughly half of the time to a new regime where it represents roughly a quarter of the corpus. By itself, this evidence might be a little less conclusive than it looks, because HathiTrust contains relatively few volumes of fiction published in the eighteenth century. Though our fiction workset contains 32,209 volumes, only 568 of them were published before 1800, and within that group several popular works (e.g., \emph{Tom Jones}) are reprinted more than ten times. But the trend in Fig. 2 can be corroborated by other sources. We have a separate corpus of 774 eighteenth- and nineteenth-century works of fiction, assembled from TCP-ECCO, the Brown Women Writers Project, and the Internet Archive. This is a significantly different kind of corpus: lacking reprints, it strives to represent diversity rather than reflect popularity. Moreover, these works were identified as fiction by scholars rather than extracted algorithmically from a larger collection. But it displays much the same shift (Fig. 3): first-person narration declines from roughly half the corpus in the eighteenth century to less than a quarter in the nineteenth. (In this corpus, first-person narration is even rarer in the nineteenth century, perhaps because eighteenth-century reprints have been excluded.) The metadata for both worksets, including classifier predictions, is available online [13].

\begin{figure}[!t]
\centering
\includegraphics[width=\linewidth]{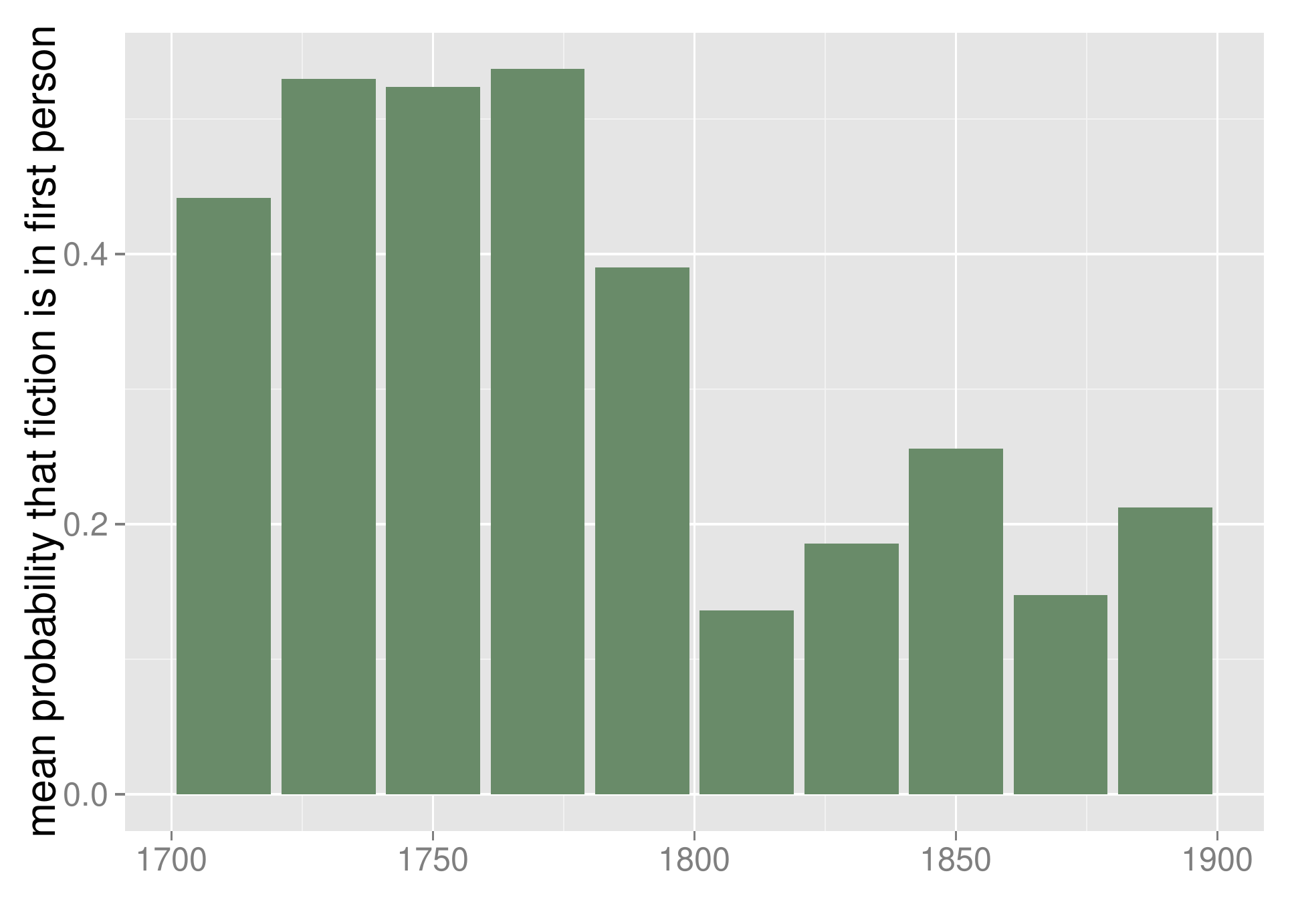}
\caption{Mean probability that fiction is written in first person, 1700-1899. Based on a corpus of 774 volumes of fiction selected by multiple hands from multiple sources. Plotted in 20-year bins because n is small here.}
\label{fig_class}
\end{figure}

Explaining the decline of first-person perspective is probably too large a task for a single article: this is a new piece of evidence for literary scholars, and it may take a few years for us to process it. But we do already know that eighteenth-century novelists enjoyed formal experiments that allowed fiction to masquerade as a real journal or autobiography or collection of letters. According to Ian Watt's timeworn but durable thesis, the novel was in fact distinguished from earlier forms of fiction by this pretense of documenting arbitrary slices of individual experience [14]. And of course novels imitating autobiography and correspondence would need to rely heavily on first-person perspective. The change at the end of the eighteenth century is harder to explain.  But there's some consensus that this period saw significant advances in the management of third-person narration, arguably culminating in Jane Austen's use of free indirect discourse, which allows readers to see through a character's eyes while retaining a third-person narrator's distanced judgment. These technical advances might have made third-person narration flexible enough that it could become a default norm in the nineteenth century. But this is speculative: there's room here for much more discussion.

\begin{table}[!t]
\renewcommand{\arraystretch}{1.3}
\caption[caption]{Features that characterized points of view \hspace{\textwidth} in a training set of 288 volumes}
\label{table_example}
\centering
% This table uses multicolumn and the booktabs package.
\begin{tabular}{ll|ll}
\toprule
\multicolumn{2}{c}{Third Person} & \multicolumn{2}{c}{First person}\\
\hline
1. his & 11. youth & 1. i & 11. got \\
2. himself & 12. \textbf{lover} & 2. we & 12. ship \\
3. herself & 13. whom & 3. us & 13. running \\
4. her & 14. silence & 4. our & 14. \textbf{eight} \\
5. he & 15. declared & 5. me & 15. another \\
6. she & 16. speech & 6. myself & 16. get \\
7. \textbf{daughter} & 17. beauty & 7. ourselves & 17. \textbf{two} \\
8. him & 18. who & 8. mine & 18. weather \\
9. whose & 19. woman & 9. \textbf{four} & 19. water \\
10. \textbf{husband} & 20. \textbf{marriage} & 10. \textbf{three} & 20. \textbf{twenty} \\
\bottomrule
\end{tabular}
\end{table}

The lists of features in Table I are almost as interesting as the historical trend itself. Although the most important features for distinguishing points of view are, predictably, personal pronouns, the Wilcoxon test also identified other themes as distinguishing first- and third-person narration. Family relationships (\emph{daughter, husband, marriage}) were prominent in third-person novels. Numbers and (perhaps) nautical terms (\emph{ship, water, weather}) were prominent in the first person.

\begin{table}[!t]
\renewcommand{\arraystretch}{1.3}
\caption{Tenfold crossvalidation of model}
\label{table_example}
\centering
% This table uses multicolumn and the booktabs package.
\begin{tabular}{l|c|c}
\hline
Actual & \multicolumn{2}{c}{Predicted}\\
\hline
  & Third person & First person \\
\hline
Third person & 165 & 4 \\
First person & 4 & 115 \\
\hline
\multicolumn{3}{c}{For first person, recall, precision, and F1 are all 0.966.}\\
\hline
\end{tabular}
\end{table}

On seeing evidence like this our first thought is always that it's an error---for instance, a quirk caused by over-representation of sea stories in our training set. But the same associations visible in Table I recur if you simply mine correlations in the whole corpus of 32,209 fiction volumes, by sorting the top 5000 words according to their degree of correlation with the ratio of first- to third-person pronouns. Even in that much larger set of works, first-person point of view correlates significantly with language of quantification. \emph{Quantity} and \emph{three} are the only examples visible at the top of Table III, but other number words also correlate positively and fell just below the cutoff for inclusion here. (\emph{Four, two, six, eight, five, several,} and \emph{twenty} are all above r = 0.15---not a huge correlation in absolute terms but certainly significant in a dataset where n = 32,209.) Likewise, although \emph{daughter} is the only kinship term visible in Table III, \emph{child, husband,} and \emph{lover} also have significant negative correlations with the first person, and are not far behind in the list. So these patterns aren't quirks caused by the contours of our training set.

Here we were conscious of the potential problem of historical heterogeneity described earlier: although the association of first-person narrative with numbers, and third-person with domesticity, is visible in the entire corpus, it might actually be caused by segregation of the corpus in some particular period. So we divided the corpus into eighteenth- and nineteenth-century halves and mined correlations in each half. But the same pattern recurred in each half of the corpus: first-person pronouns are strongly associated with numbers, and third-person pronouns with terms of kinship.

\begin{table}[!t]
\renewcommand{\arraystretch}{1.3}
\caption[caption]{Strongest correlations with the ratio \hspace{\textwidth} first-person pronouns / third-person pronouns \hspace{\textwidth} in a corpus of 32,209 fiction volumes,\hspace{\textwidth} excluding pronouns and the verb \emph{to be}.}
\label{table_example}
\centering
% This table uses multicolumn and the booktabs package.
\begin{tabular}{ll|ll}
\toprule
\multicolumn{2}{c}{Negative correlations} & \multicolumn{2}{c}{Positive correlations}\\
\hline
eyes & -0.214 & tonne & 0.308 \\
voice & -0.188 & shore & 0.285  \\
face & -0.181 & pieces & 0.265  \\
lips & -0.177 & \textbf{quantity} & 0.250  \\
smile & -0.166 & provisions & 0.241 \\
girl & -0.165 & \textbf{three} & 0.239 \\
glance & -0.159 & northeast & 0.238  \\
\bf{daughter} & -0.158 & island & 0.237  \\
silent & -0.158 & voyage & 0.237 \\
pale & -0.157 & southeast & 0.218 \\
turned & -0.156 & habitation & 0.213 \\
trembling & -0.151 & ship & 0.211\\
loved & -0.149 & boat & 0.209\\
woman & -0.149 & powder & 0.207\\
watched & -0.148 & piece & 0.207\\
\bottomrule
\end{tabular}
\end{table}

Of course there are several other striking facts about the correlations in Table III. In the list of negative correlations, body parts and verbs associated with emotion are even more salient than kinship words. There's also a clear emphasis on feminine gender (\emph{girl, daughter, woman}). It's possible that women wrote relatively more often in the third person, but also possible that these patterns are explained by reliance on third-person narration in a particular genre (like courtship fiction). We plan to test both hypotheses soon.

On the other hand, first-person pronouns are positively correlated with terms that describe travel---and especially travel by sea. We found one clue to this association when we looked at a list of titles sorted by our classifier. A remarkable number of the works recognized as most strongly ``first-person'' in character belonged to the genre of the Robinsonade. We were aware that \emph{Robinson Crusoe} (1719) was a widely-reprinted and frequently-imitated novel, but we hadn't realized just how long and vigorously this tradition endured. The \emph{Swiss Family Robinson} (1812) was joined by an \emph{English Family Robinson} (1852), for instance, and eventually by a \emph{Robinson Crusoe of the Nineteenth Century} (1884), and many others.

The castaway premise obviously encourages a story to rely on first-person narration, and we suspect it may also encourage quantification. Crusoe's obsession with enumeration and accumulation is famously part of the colonial logic of his narrative, which transforms (ostensibly) uninhabited land into a well-governed productive system. \begin{quote}
[I]n about a Year and a Half I had a flock of about twelve Goats, Kids, and all; and in two Years more I had three and forty, besides several that I took and kill'd for my Food. After that, I enclosed five several Pieces of Ground to feed them in, with little Penns to drive them into, to take them as I wanted, and Gates out of one Piece of Ground into another [15].
\end{quote} But the genre of the Robinsonade by itself wouldn't explain the strength of a correlation that holds across the whole corpus of eighteenth- and nineteenth-century fiction. We suspect that the combination of a first-person narrator, a colonial setting, and acquisitive quantification extends beyond the Robinsonade to encompass other kinds of adventure fiction---from Òimperial romancesÓ like \emph{King Solomon's Mines} (1885) to sea-stories and westerns and perhaps even stranger forms of ethnographic travel like Wells' \emph{Time Machine} (1895). But this hypothesis needs to be confirmed by further investigation.

\subsection{Conclusion: the historical value of classification.}

We began by presenting classification as a practical support system for humanistic data mining---dividing digital collections into known generic categories in order to support further inquiry.

Classification algorithms can in fact support an initial map of this kind. They can supplement existing metadata in cases where it's patchy or insufficiently detailed, and they can help us grapple with problems that involve differences of degree rather than clear categorical boundaries.

But classification algorithms can also become a method of conceptual exploration in their own right. Often we associate exploratory analysis with unsupervised methods like topic modeling. Supervised learning  may seem pedestrian by contrast, since supervised algorithms can only answer questions we already know how to pose. But knowing how to pose a question doesn't mean you'll get the answer you expect! You may after all discover that your question was badly formed. Genre is not a well-defined concept, and in attempting to map it digitally, we're beginning to discover that it breaks up into a range of different categories.

For instance, even the initial explorations described here have revealed genres of four different kinds. 1) At the broadest level, concepts like prose fiction and lyric poetry are relatively stable across a timespan of centuries, and relatively easy to map using the evidence of diction. Then there are 2) generic phenomena of a briefer kind, but almost equally easy to map using the evidence of diction. We've trained a model that recognizes Romantic-era gothic novels. Matthew Jockers has done the same thing for many other subgenres of nineteenth-century fiction, and he confirms Franco Moretti's speculation that these genres often seem to have a roughly generational (30-year) duration [16].

But the tradition of literary scholarship we inherit also describes 3) specific subgenres of fiction, drama, and poetry that endure a century or longer, like ``the gothic novel'' (broadly construed) or ``science fiction." Some of these concepts may turn out to be difficult to recognize lexically: for instance, we haven't been able to train a single, unified model of the gothic across the whole nineteenth century. But of course the failure of our algorithm doesn't imply that existing concepts of the gothic lack coherence. There's no reason to assume that all genres must be recognizable at the level of diction.

Finally, in some cases we may discover 4) long-lived subgenres characterized by a surprising kind of lexical coherence. The two-century-long tradition of the Robinsonade is one example; it may even reveal an undescribed dimension of narrative that unites previously distinct genres of eighteenth- and nineteenth-century fiction through a shared emphasis on enumeration. But this hypothesis needs more investigation. It's apparent that genre and point of view aren't independent axes of description, but the nature of the relationship between those axes hasn't yet been described adequately.

\section{Internal heterogeneity}

One limitation of the research described above is that we have treated volumes as if they were unified entities. In reality, the volumes in a digital library are almost always internally heterogeneous. In the most extreme case, the collected works of an author may mix poetry and drama with personal letters and prose essays. Nineteenth-century periodicals are equally miscellaneous. But nearly every volume in a library poses some version of this problem, because bookplates, indexes, and publishers' ads don't belong to the same genre as the body text itself. Since that paratext represents 10\% or more of the text in many volumes, it's a problem that really needs to be addressed.

The Text Encoding Initiative represents an ideal solution to this problem for small and medium-sized collections, but it's not immediately practical to manually mark up the millions of volumes in HathiTrust. If document formats were consistent, we might devise segmentation procedures that mimicked a human reader by looking for changes in font size, or even parsing the grammar of a title page [17]. But in a library that covers several centuries, the protocols of book layout are anything but consistent. Moreover, optically-transcribed documents pose a problem for algorithms that tolerate noise poorly. Our OCR correction procedures catch common errors (like eighteenth-century s$\rightarrow$f substitution) that would systematically distort our findings [18]. But it's not safe to assume that any particular line will be transcribed correctly.

\subsection{Proposed solution.}

We've accordingly opted for a more robust strategy that uses classification itself as a segmentation tool. We don't, after all, need to separate the parts of a volume unless they're generically distinct. It may ultimately be desirable to separate the individual chapters of a novel or individual poems in a collection, but our immediate goal is the simpler one of locating page ranges that contain poetry, and separating them from pages that contain fiction. This problem can be addressed most directly by training classification algorithms to recognize the genres in question.

We do this training initially at the page level. The parts of a volume don't always begin on a new page, but divisions within a page can be more easily addressed after generically coherent page ranges have been identified.

To treat pages as separate documents would, however, discard the valuable evidence provided by sequence. The sequential arrangement of pages provides clues of several kinds. Most obviously, there are parts of a volume (like a bookplate or an index) that tend to be located at the beginning or end. More importantly, sequence helps us smooth out noise; in the middle of a long sequence of fiction pages, a page identified as nonfiction is likely to be an error. The relative position of volume parts can also be significant: biography is relatively difficult to distinguish from fiction at the level of the individual page, but biographical introductions often precede a table of contents, and fiction rarely does.

\begin{figure}[!t]
\centering
\includegraphics[width=\linewidth]{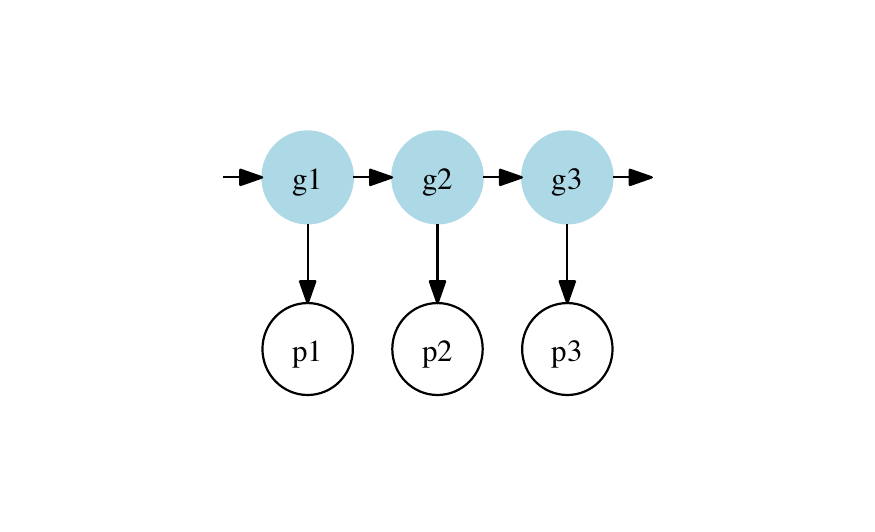}
\caption{Hidden Markov model. The hidden states (genre1, genre2, ...) are inferred from observed states (page1, page2, ...) and from a transition matrix defining the probabilities of transition between different hidden states.}
\label{fig_sim}
\end{figure}

One flexible way to represent this kind of knowledge about volume structure might be to train a hidden Markov model on top of the predictions made by page-level classifiers. In other words, we model volume structure as a sequence where the genre of each new page depends stochastically on the genre of the page that preceded it. We train this model on manually labeled ground truth, so it can learn, for instance, that a page of fiction is very likely to be followed by another page of fiction, less likely to be followed by an index, and not at all likely to be followed by a table of contents. However, this is a \emph{hidden} Markov model, because the hidden states (genres) are never directly observable outside of the training environment [19]. Instead, twenty classifiers make predictions about the probability that a page instantiates each of twenty genres. The hidden Markov model defines a principled smoothing procedure whereby we combine those predictions about individual pages with our knowledge about the probabilities of transition between genres to infer the most likely hidden state (genre) for every page in the sequence.

\subsection {Results.}

How much does this technique improve our results? 

First, we should characterize the accuracy of page-level classification without Markov smoothing. We trained page-level classifiers using the Weka implementation of regularized logistic regression (L2 norm) because it's a robust algorithm for text classification that expresses predictions probabilistically [20]. Naive Bayes and SVM might make equally accurate decisions, but expressing their predictions as probabilities of genre membership would require an additional calibration step.

We trained classifiers for individual classes, and when not using smoothing we identified the genre of a page simply by accepting the prediction with the highest probability (the one-vs-all method). We also used a practical trick well attested in industrial practice [21], and given a theoretical basis by Nigam, McCallum, and Mitchell [22]: when a class showed internal variation, we identified it through an ensemble of classifiers designed to recognize its subclasses. For instance, nonfiction prose is a large category, containing genres like ``biography'' and ``autobiography'' that are especially difficult to distinguish from fiction. We trained classifiers for all of these subclasses, and if \textit{any one} of them turned out to be the most probable genre for a given page, that page was identified as nonfiction. Similarly, we divided front matter into subclasses like ``title page'' and ``bookplate,'' and back matter into subclasses like ``index" and ``date due slip."

Regularized logistic regression makes feature selection less crucial, since it minimizes features that turn out not to be useful for discriminating a particular class, but resists overfitting [21, 23]. We empirically optimized the number of features, and ended up using the 250 most common words in the corpus (including a few word categories compressed into single features, like personal-name and roman-numeral). We also used information about the relative length of pages, their position in the volume, and the density of line-initial capitalization.

\begin{figure}[!t]
\centering
\includegraphics[width=\linewidth]{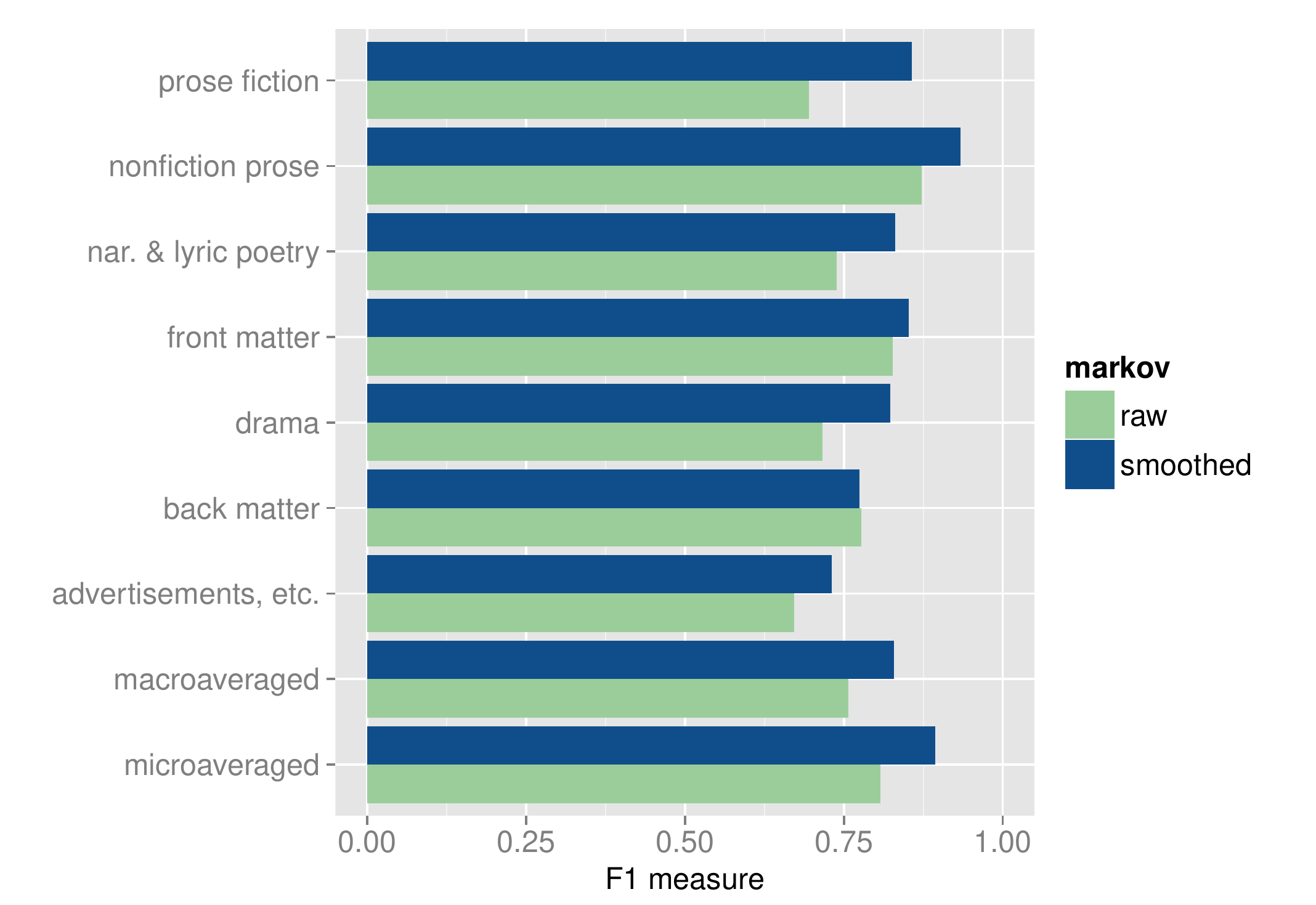}
\caption{Gains from hidden Markov smoothing. The top seven rows are F1 measures for individual genres; the bottom two rows reflect macro- and micro-averaged F1 measures for all genres.}
\label{fig_sim}
\end{figure}

We manually recorded genre labels for 101 volumes (31,586 pages) drawn randomly from 469,200 eighteenth- and nineteenth-century volumes, and performed classification using tenfold cross-validation, with the corpus divided by volume rather than page. (Dividing the corpus by page would make the experimental task much easier than it is in reality, since you would be comparing authors to themselves; division by volume is also necessary in order to make an apples-to-apples comparison with the smoothed results.) The F1 scores for different genres are presented in the ``raw" bars of Fig. 5. (The F1 metric is important in this context because the classes are very different in size, and it would be easy to maximize a crude accuracy metric by wholly ignoring some of the smaller classes.)

Then, using the same tenfold division of the corpus, we smoothed the classifiers' predictions using a different hidden Markov model for each iteration of cross-validation. (The model was never allowed to see the test set.) The smoothed results are presented in the ``smoothed'' bars of Fig. 5, and show substantial improvement. Most of the improvement takes place in the larger categories that comprise the majority of pages in a volume (nonfiction prose, fiction, etc.), and it seems to be produced mostly by smoothing out minor variations within long generically-coherent sequences of pages.

\subsection{Remaining challenges and scalability.}

It's fairly clear that Markov smoothing can be a useful tool for segmenting heterogeneous volumes, but also clear that it will need to be supplemented by other strategies. F1 scores in the range of .80 -- .90 are good, as algorithms go, but humanists have a lower tolerance for error.

As we adapt this technique for practical use, we can improve it in several ways. First, 100 volumes is actually a tiny training set given the diversity of this corpus, and it should be enlarged manually. This technique would also be well suited to a semi-supervised bootstrap approach---where the algorithm is allowed to generate additional data for its own further training. Because there are actually two different models involved here (a volume-structure model and a bag-of-words model) they should tend to correct each other and prevent cascading error, in a system known as ``co-training'' [24].

It may also be possible to improve our model. The sequence of pages in a volume isn't actually a first-order Markovian sequence; rather, it tends to have a hierarchical character governed by large structural divisions between front matter, preface, body text, and back matter. A hierarchical hidden Markov model might be appropriate, or we could use conditional random fields.

We'll also want to refine our techniques to make divisions below the page level. This will require two different strategies. In some cases, a volume part simply begins halfway down a page. Once we've identified coherent page ranges, it should be possible to catch most cases like this by defining a similarity metric for each range and scanning the pages on either side of the range boundary for points of inflection. An XML tag can then be inserted at an appropriate location. Small errors here are not likely to affect macroscopic results. A more important problem occurs where two genres are mixed throughout a page range. For instance, nineteenth-century poems often have voluminous prose footnotes on every page.

Some generic mosaics of this kind will be difficult to divide algorithmically. But we've found that the most common and important case---a mosaic that mixes prose and verse---is fairly straightforward. In the eighteenth and nineteenth centuries, verse is easily distinguished from prose, on a line-by-line level, using a combination of line length and initial capitalization. This would be difficult in the twentieth century, since verse isn't always capitalized. But since twentieth-century poems don't usually have long prose footnotes, the solution is in practice tailored to the problem.

In this article we've focused on scale mainly as a conceptual problem (involving historical timeframes and heterogeneous volumes). But it's also, of course, a processing problem. Compressed, the dataset described in this article amounts to 0.7TB; the processes we describe are implemented in Python and Java, and have been parallelized in straightforward ways (we write multithreaded Java, and use multiple nodes in our campus cluster). Although we expect the size of the problem to double or triple as we expand the project, it remains manageable for researchers comfortable with batch queues.

But scalability still poses a social problem, because most humanists are not in fact comfortable with batch queues. If we want our methods to be adopted by other scholars, we'll need to share them in a form that can be managed more easily on a personal workstation. The most realistic path to this goal is to integrate our analytical methods in the digital library that hosts the original data. For that reason we're collaborating with HathiTrust Research Center to implement our methods in their online research portal (this project is supported by the National Endowment for the Humanities and the American Council of Learned Societies).

\section{Conclusion: Integrating both dimensions of this problem.}

This article has described two different problems of scale: one that involves segmenting individual volumes, and one that involves mapping historically mutable genres. To solve both problems at once, we'll need to define relationships between different levels of classification. At the page level, it's not necessary to distinguish subgenres like the gothic novel  or science fiction. Subgenres of that kind aren't very useful for segmenting volumes, because it rarely makes sense to treat them as mutually exclusive categories (a novel can be at once gothic and science fiction, as Mary Shelley would remind us). Instead, page-level classification will focus on relatively broad categories like prose fiction, nonfiction, drama, and nondramatic poetry. These categories can generally be positioned in a zero-sum relationship to each other, and at the page level it's not too much of a distortion to choose one of them in exclusion to the others. Once we've separated these categories within volumes, we can treat cohesive page ranges as ``works" for subtler multi-label classification at a historical scale, where unlimited numbers of subgenres can be allowed to overlap in a single work.

This multilayered approach doesn't, however, imply that the problem of volume segmentation can be treated entirely ahistorically. Volume structure does vary across time: eighteenth-century volumes tend to be prefaced by lists of subscribers, whereas nineteenth-century volumes end with advertisements. So we'll probably need to train multiple Markov models of volume structure on different spans of time. 

If patient readers are at this point dismayed by the complexity of the problem, we sympathize. Humanistic data mining is immensely rewarding, because it enlarges our field of vision in a way that can lead to significant discoveries. But there are many potential sources of error in this domain, from data-cleaning problems to hasty assumptions about historical continuity. A principled method is necessary here because researchers will otherwise be overwhelmed by innumerable ad hoc patches addressing specific kinds of uncertainty.

We understand machine learning, not as a black box that produces authoritative results, but as a flexible way to explore large collections that incorporates reflection on uncertainty as an integral part of its logic. The Bayesian concepts underpinning this discipline seem to dovetail rather well with humanistic insights like historicism. Ideally, humanists will approach machine learning, not just as a collection of tools, but as a principled statistical language that allows them to formalize and scale up their own habits of inquiry.

\section*{Acknowledgments}
Work on this project was supported by the Andrew W. Mellon Foundation and by NEH Digital Humanities Start-Up Grant \#HD-51787-13. The findings and recommendations expressed here do not necessarily reflect the views of those granting agencies. The main collection we used was drawn from HathiTrust Digital Library with assistance from HathiTrust Research Center. Corroborating evidence was drawn from a smaller collection assembled with help from Jordan Sellers. Early drafts of the first half of this project were presented at the Nebraska Forum on Digital Humanities and the Stanford Literary Lab, and online at \emph{The Stone and the Shell}, where feedback came from many participants and commenters. David Bamman, Sayan Bhattacharyya, Tim Cole, Eleanor Courtemanche, M. J. Han, William Underwood, and Laura White gave particularly valuable advice.

\end{document}